\documentclass[journal]{IEEEtai}
\usepackage[colorlinks,urlcolor=blue,linkcolor=blue,citecolor=blue]{hyperref}
\usepackage{color,array}
\usepackage{lipsum}
\usepackage[utf8]{inputenc} 
\usepackage[T1]{fontenc}    
\usepackage{url}  
\usepackage{hyperref}       
\hypersetup{
    colorlinks=true,       
    linkcolor=blue,          
    citecolor=blue,        
    urlcolor=blue           
}
\usepackage{booktabs}       
\usepackage{nicefrac}       
\usepackage{microtype}      
\usepackage{lipsum}		
\usepackage{graphicx}
\usepackage{doi}
\usepackage{amsmath}
\usepackage{subfig}
\usepackage[affil-it]{authblk}
\usepackage[english]{babel}
\usepackage{blindtext}
\usepackage{amssymb}
\usepackage{cleveref}
\usepackage{booktabs}
\usepackage{framed,multirow}
\usepackage{hyperref}
\usepackage{url}            
\usepackage{booktabs}
\usepackage[mathlines]{lineno}
\usepackage{amsfonts}       
\usepackage{latexsym}
\usepackage{amsmath}
\usepackage{amssymb}
\usepackage{amsfonts}

\usepackage{cleveref}
\usepackage{nicefrac}       
\usepackage{microtype}      
\usepackage{lipsum}
\usepackage{graphicx}
\usepackage{adjustbox}
\usepackage{graphicx}
\usepackage[table,xcdraw]{xcolor}
\definecolor{newcolor}{rgb}{.8,.349,.1}
\usepackage[linesnumbered,ruled,vlined]{algorithm2e}

\usepackage{multirow}
\usepackage{diagbox}
\usepackage{subfig}
\usepackage{tikz}
\usetikzlibrary{external}
\newtheorem{definition}{Definition}
\usepackage{pgfplots}
\pgfplotsset{compat=1.7}
\usetikzlibrary{positioning}
\usepackage{etoolbox} 
\usepackage[numbers,sort]{natbib}

\newcommand*\linenomathpatch[1]{%
  \cspreto{#1}{\linenomath}%
  \cspreto{#1*}{\linenomath}%
  \csappto{end#1}{\endlinenomath}%
  \csappto{end#1*}{\endlinenomath}%
}
\newcommand*\linenomathpatchAMS[1]{%
  \cspreto{#1}{\linenomathAMS}%
  \cspreto{#1*}{\linenomathAMS}%
  \csappto{end#1}{\endlinenomath}%
  \csappto{end#1*}{\endlinenomath}%
}
\expandafter\ifx\linenomath\linenomathWithnumbers
  \let\linenomathAMS\linenomathWithnumbers
  \patchcmd\linenomathAMS{\advance\postdisplaypenalty\linenopenalty}{}{}{}
\else
  \let\linenomathAMS\linenomathNonumbers
\fi

\linenomathpatch{equation}
\linenomathpatchAMS{gather}
\linenomathpatchAMS{multline}
\linenomathpatchAMS{align}
\linenomathpatchAMS{alignat}
\linenomathpatchAMS{flalign}

\SetCommentSty{mycommfont}

\crefformat{section}{\S#2#1#3}
\crefformat{subsection}{\S#2#1#3}
\crefformat{subsubsection}{\S#2#1#3}
\crefrangeformat{section}{\S\S#3#1#4 to~#5#2#6}
\crefmultiformat{section}{\S\S#2#1#3}{ and~#2#1#3}{, #2#1#3}{ and~#2#1#3}

\begin{document}

\title{Characterizing and Mitigating the Difficulty in Training Physics-informed Artificial Neural Networks under Pointwise Constraints}

\author{\href{https://orcid.org/0000-0002-1095-0881}{\includegraphics[scale=0.08]{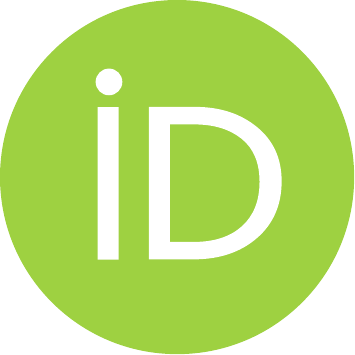}\hspace{1mm}Shamsulhaq Basir}, \href{https://orcid.org/0000-0003-1967-7583}{\includegraphics[scale=0.08]{orcid.pdf}\hspace{1mm}Inanc Senocak}

\thanks{This material is based upon work supported by the National Science Foundation under Grant No. 1953204 and in part by the University of Pittsburgh Center for Research Computing through the resources provided.}

\thanks{Shamsulhaq Basir, is a PhD candidate in the Mechanical Engineering and Materials Science Department at the University of Pittsburgh, Pittsburgh, PA 15261, USA. (e-mail:shb105@pitt.edu)}

\thanks{Inanc Senocak, is an associate professor in the Mechanical Engineering and Materials Science Department at the University of Pittsburgh, Pittsburgh, PA 15261, USA. (e-mail:senocak@pitt.edu)}}

\maketitle
\begin{abstract}
Neural networks can be used to learn the solution of partial differential equations (PDEs) on arbitrary domains without requiring a computational mesh. Common approaches integrate differential operators in training neural networks using a structured loss function. The most common training algorithm for neural networks is backpropagation which relies on the gradient of the loss function with respect to the parameters of the network. In this work, we characterize the difficulty of training neural networks on physics by investigating the impact of differential operators in corrupting the back propagated gradients. Particularly, we show that perturbations present in the output of a neural network model during early stages of training lead to higher levels of noise in a structured loss function that is composed of high-order differential operators. These perturbations consequently corrupt the back-propagated gradients and impede convergence. We mitigate this issue by introducing auxiliary flux parameters to obtain a system of first-order differential equations. We formulate a non-linear unconstrained optimization problem using the augmented Lagrangian method that properly constrains the boundary conditions and adaptively focus on regions of higher gradients that are difficult to learn. We apply our approach to learn the solution of various benchmark PDE problems and demonstrate orders of magnitude improvement over existing approaches.   
\end{abstract}

\begin{IEEEImpStatement}
In the field of physics-informed machine learning, neural networks can be used to efficiently learn physical phenomena governed by differential equations. Here, we identify a key difficulty in training neural networks when governing equations contain high order differential operators. We analyze the associated learning complexity issue and propose an approach that can efficiently tackle problems that have been challenging to learn with existing neural network methods. 
\end{IEEEImpStatement}

\begin{IEEEkeywords}
Constrained optimization, Augmented Lagrangian method, meshless method, machine learning
\end{IEEEkeywords}

\section{Introduction}\label{sec:introduction}
Partial differential equations (PDEs) play a vital role in our comprehension of a wide range of physical phenomena, including  sound propagation, heat and mass transfer, fluid flow, and elasticity to name a few. Most of the modern problems involving PDEs are usually solved via numerical methods, owing to the lack of closed analytical solutions. The most common and powerful numerical methods for solving PDEs are finite volume, finite difference, finite element, and spectral element methods. Although these methods are highly efficient in solving forward problems, they do not readily extend to data-driven modeling and inverse problems. Furthermore, quality mesh generation is an essential part of conventional numerical methods, which can be tedious and time consuming for problems involving complex geometry. To this end, neural networks as universal approximators \cite{hornik1989multilayer} can be viewed as an alternative meshless approach to solve PDEs in either a strong or weak form by randomly distributing points within the solution domain.

\citet{dissanayake1994neural} and \citet{van1995neural} are credited with introducing neural networks as an alternative solution technique for PDEs.  Their pioneering approach with neural networks have been applied to learn the solution of different types of PDEs with satisfactory results \cite{monterola2001solving,quito2001solving,Parisi2003solving,hayati2007feedforward}. Different from those previous works, \citet{lagaris1998artificial} proposed a neural network-based method for the solution of differential equations on orthogonal box domains. Their approach relies on constructing custom trial functions that satisfy boundary conditions by construction. However, their approach is limited to simple domains for which it is trivial to create trial functions. In addition, creating trial functions impose prior bias toward a certain class of functions that might not be optimal for the problem at hand. Most recently, several researchers have taken a similar approach to apply neural networks for the solution of differential equations either in the weak or strong form \cite{weinan2017proposal,Raissi2019,sirignano2018dgm}. \citet{weinan2017proposal} proposed the Deep Ritz method for the solution of PDEs. However, their method is only applicable to problems that can be formulated as energy minimization problems. \citet{sirignano2018dgm} proposed Deep Galerkin method (DGM) for the solution of high-dimensional PDEs. Similar to the early works in \cite{dissanayake1994neural,van1995neural,Parisi2003solving}, \citet{Raissi2019} proposed physics-informed neural networks (PINNs) using a modern deep-learning framework TensorFlow \cite{abadi2016tensorflow}. The interest to use neural networks to learn the solution of PDEs continues to grow at a fast pace with applications in various domains \cite{raissi2019deep,kissas2020machine,mao2020physics,gao2021phygeonet,ALIAKBARI2022109002,PATEL2022110754,AMININIAKI2021113959,Chen20,ALMAJID2022109205,GAO2021110079}. In the present work, we refer to the technical approach pursued in \cite{dissanayake1994neural,van1995neural,monterola2001solving,Parisi2003solving,hayati2007feedforward,Raissi2019} as PINNs because the core formulation in these works are essentially the same.

A common technique pursued in PINNs is to minimize a weighted sum of several objective functions that are balanced with multiplicative weighting coefficients or hyperparameters. These hyperparameters are not known \textit{a priori}.  Many researchers have found that predictions from neural network models are highly dependent on these hyperparameters and \textit{a priori} determination of them have been a research topic \cite{van2020optimally,mcclenny2020self,wang2021understanding,krishnapriyan2021characterizing,basir2022critical,PECANN_2022,bischof2021multi}. \citet{wang2021understanding} demonstrated that PINNs do not produce consistent and physically feasible solutions when applied to various kinds of PDEs. \citeauthor{wang2021understanding} proposed an empirical algorithm that improves over the conventional PINNs, but even their method has several limitations as we have discussed in our prior work \cite{PECANN_2022}. \citet{van2020optimally} proposed a heuristic method to determine these hyperparameters by considering an affine combination of a physics-informed bi-objective loss function. However, their approach also has issues. First, it is not advisable to sum up objective functions with different scales to form a mono-objective optimization equation \cite{lobato2017multi}, because the objective function representing the residual on a given PDE and the one representing the mismatch on boundary conditions do not often share the same scale and, therefore, the learned hyperparameters no longer represent the relative importance of the objective functions. Second, for non-convex Pareto fronts, some optimal set of solutions cannot be found with any combination of the weighting factors \cite{lobato2017multi}. In a different work \cite{basir2022critical}, we visualized the loss landscapes of a trained data-driven neural network model and its physics-informed counterpart and demonstrated that incomparable scales between loss terms in a composite objective function impede the convergence of neural network models.

Recently, we proposed physics and equality constrained artificial neural networks (PECANNs) to learn the solution of forward and inverse problems \cite{PECANN_2022}. Our PECANN framework is  noise-aware and adept at multi-fidelity data fusion. The backbone of the PECANN framework is a constrained optimization formulation that is recast as an unconstrained optimization problem  using the augmented Lagrangian method \cite{powell1969method, bertsekas1976multiplier}. The PECANN framework balances each term in the objective function in a principled fashion and enable the user to specify the degree of noise in observed data. It is worth noting that in the PINN approach and as well as in the PECANN approach, $L^2$ norm is used in training the neural network. However, $L^2$ norm increases the learning complexity of the original problem \cite{cai2020deep}, which is further elevated for PDE solutions that are ill-conditioned. Because the predictions of the network are often noisy or incorrect during the early stages of training, noise in the back-propagated gradients can be amplified and impede the convergence of PDE problems with ill-conditioned solutions. The learning complexity can be mitigated by reducing the order of the differential operators. For instance, \citet{weinan2017proposal} uses a variational formulation to reduce the order of a differential operator via integration by parts. However, their approach can be applied only to problems that can be formulated as an energy minimization problem. In addition, boundary conditions are soft-constrained in their approach by a penalty parameter that are not known \textit{ a priori}. \citet{cai2020deep} proposed a method that adopts least-squares functionals to train a deep neural network to learn the solution of one-dimensional elliptic PDEs. The least-squares functionals are based on the so-called first-order system least-squares (FOSLS). Through this approach, the authors are able to reduce the order of a given elliptic PDE by introducing an auxiliary flux parameter. Their method learns the primary variable and the auxiliary flux variable by a composite neural network with two branches. Since Sobolov norms (i.e. $\|\cdot\|_{1/2}$ or $\|\cdot\|_{-1/2}$) are computationally infeasible, the authors approximated them with weighted $L^2$ norms. Boundary conditions on the PDE are enforced as penalty regularizers with hyperparameters that are known in \textit{a priori}. Differential operators are approximated using a finite difference scheme on a fixed mesh. However, requiring a computational mesh chips away the appeal of using neural networks as a meshless method. Also, finite difference approximation of derivatives requires multiple forward passes through the computational graph amplifying the inherent discretization errors, particularly for problems with multiple inputs. Whereas with  automatic differentiation, which computes the derivatives in a single backward pass, derivatives can be calculated at machine precision  \cite{baydin2018automatic}.

In the present work, we demonstrate how structured objective functions that contains differential operators amplify noise in the learning process, which in turn corrupt the back propagated gradients. Consequently, the convergence of the learning process is impeded. This issue is amplified for PDE solutions that are inherently ill-conditioned. We then propose a meshless neural network-based methods for PDEs that may not have any underlying energy minimization principles. We achieve this by introducing auxiliary flux parameters to obtain a first order system of equations, which can also be viewed as preconditioning a given PDE. This preconditioning precludes the approximation of higher derivatives and therefore mitigate the learning complexity of the problem by relaxing the stringent smoothness requirement of the solution. We formulate an unconstrained optimization problem that properly enforces the boundary conditions and allows for an adaptive attention mechanism that focuses on the regions with higher gradients that are challenging to learn. Unlike the composite neural-network approach adopted in \cite{cai2020deep}, where numerical quadrature is used to compute the first-order system, our proposed method uses a single neural network architecture and employ automatic differentiation (AD) \cite{baydin2018automatic}.

\subsection{Effect of Differential Operators in Corrupting the Back-propagated Gradients}
In this section, we demonstrate how a differential operator amplifies noise in the predicted solution and, hence, corrupt the back-propagated gradients which may impede convergence. We use a simple one-dimensional Poisson's equation to explain this issue as follows:
\begin{align}
    \frac{d^2 u(x)}{dx^2} &= f(x), \forall x \in \Omega = [0,1],\\
    u(x) &= g(x), \in \partial \Omega,
\end{align}
where $\Omega$ is the domain with its boundary $\partial \Omega$, $f(x)$ and $g(x)$ are source function and boundary function respectively. We manufacture a simple solution as $u(x) = \sin(5 \pi x)$ for the above differential equation. $g(x)$  and $f(x)$ can be calculated exactly using the manufactured solution. The objective function using the PECANN framework\cite{PECANN_2022} can be written as follows:
\begin{align}
    \mathcal{L}(\theta) &= \mathcal{L}_{\mathcal{D}}(\theta) + \sum_{i=1}^{N_{\partial \Omega}} \lambda^{(i)} \phi(\hat{u}(x^{(i)}) - g(x^{(i)})) + \frac{\mu}{2}\pi,\label{eq:ill_condition_loss}
    \\
    \mathcal{L}_{\mathcal{D}}(\theta) &= \sum_{i=1}^{N_{\Omega}} \|\frac{d^2 \hat{u}(x^{(i)})}{dx^2} - f(x^{(i)})\|_2^2,\\
    \pi &= \sum_{i=1}^{N_{\partial \Omega}} \| \phi(\hat{u}(x^{(i)}) - g(x^{(i)}))||_2^2,
    \label{eq:ill_condition_penalty}
\end{align}
where $\hat{u}$ is the prediction of our neural network model, $\lambda \in \mathbf{R}^2$ is a vector of Lagrange multipliers, $\mu$ is a positive penalty parameter and $\phi$ is a quadratic distance function. For this one dimensional problem, the number of boundary points $N_{\partial \Omega}$ is 2. We note that the objective function given in Eq.~\eqref{eq:ill_condition_loss} has a second order differential operator. 

We carry out a numerical experiment to demonstrate the impact of a second-order differential operator on corrupting the back-propagated gradients in the presence of perturbations in the predicted solution during training. In our example, we use a feed-forward neural network with two hidden layers and 20 neurons per layer. We use a Sobol sequence with $N_{\Omega} =256$ residual points uniformly in the domain as well as two boundary conditions only once before training. 

First, we train our network for 1000 epochs and predict the solution from our neural network model $u(x;\theta)$ for comparison purposes. We also record the parameters of the network (i.e., $\theta$) and their gradients (i.e., $\nabla \mathcal{L}_{\theta}$). Next, we perturb our parameters (i.e., $\theta$)  to obtain a new set of parameters (i.e., $\tilde{\theta}$). Perturbations are generated using two random Gaussian vectors with appropriate scaling as proposed in \cite{li2018visualizing}. We then make a prediction from our neural network model $u(x;\tilde{\theta})$ at the perturbed state. We then obtain the back-propagated gradients (i.e., $\nabla \mathcal{L}_{\Tilde{\theta}}$) at the perturbed state.

\begin{figure*}[!ht]
\centering
    \subfloat[]{\includegraphics[scale=0.65]{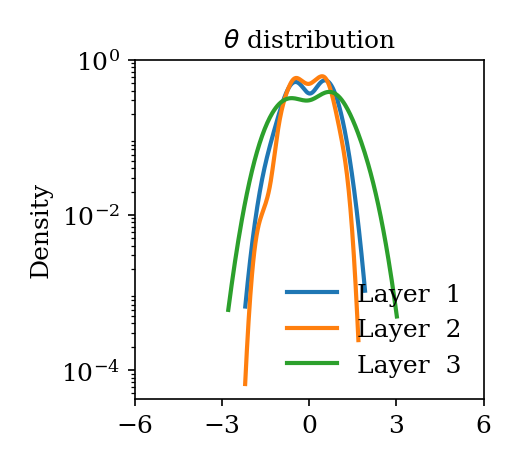}}\hspace{4em}
    \subfloat[]{\includegraphics[scale=0.65]{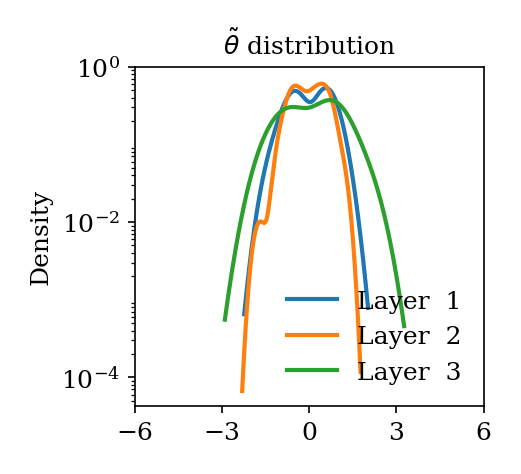}} 
    \caption{Distribution of the parameters of our neural network model at different states:  (a) at the end of training for 1000 epochs, (b) after injection of perturbations}
    \label{fig:model_params}
\end{figure*}

From Figs.~\ref{fig:model_params}(a) and (b) we observe that the distribution of the parameters of our network at the end of training and after the addition of perturbations are fairly similar, which indicates acceptable levels of noise or perturbations.

\begin{figure*}[!h]
\centering
    \subfloat[]{\includegraphics[scale=0.65]{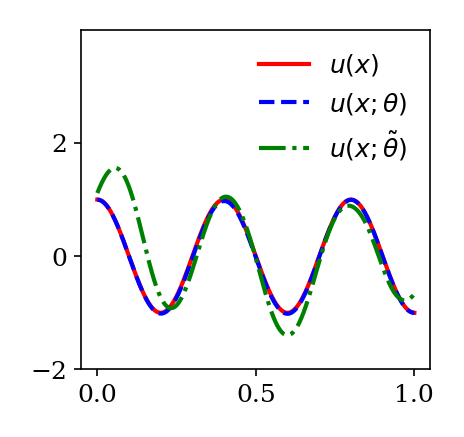}}\hspace{6em}
    \subfloat[]{\includegraphics[scale=0.65]{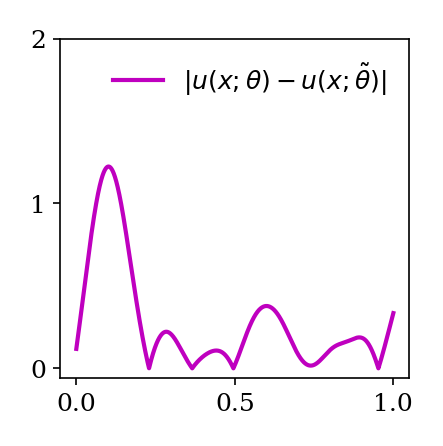}} \\
    \caption{ (a) exact solution $u(x)$, predicted solution $u(x;\theta)$ in dashed blue obtained from our neural network model after training for 1000 epochs, predicted solution $u(x;\tilde{\theta)}$ in dashed green obtained from our neural network model at the perturbed state, (b) absolute point-wise difference between $u(x;\theta)$ and $u(x;\tilde{\theta})$,}
    \label{fig:model_predictions}
\end{figure*}

Next, we investigate the impact of the perturbations on the predictions obtained from our neural network model. In Fig.~\ref{fig:model_predictions}(a), we observe that our model has produced an acceptably accurate prediction of the solution after training without any perturbations. However, after the addition of perturbations, the prediction of our model $u(x;\tilde{\theta})$ distinctly deviates from the exact solution as seen in Fig.~\ref{fig:model_predictions}(a). The error resulting from this deviation is plotted in Fig.~\ref{fig:model_predictions}(b), which quantifies the absolute point-wise difference between the prediction of our model before and after the injection of noise in the parameters $\theta$ of our neural network.

\begin{figure*}[!h]
\centering
    \subfloat[]{\includegraphics[scale=0.650]{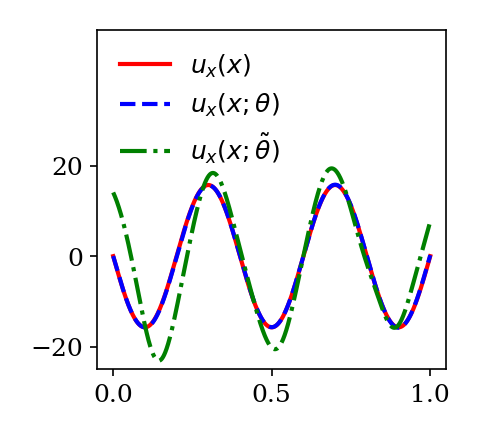}}\hspace{6em}
    \subfloat[]{\includegraphics[scale=0.650]{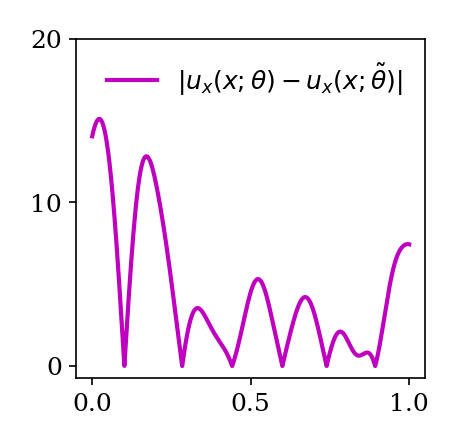}}\\
    \caption{First-order derivative of the prediction obtained from our neural network model at different states: (a) comparison of the exact first derivative $u_x(x)$, predicted first derivative $u_x(x;\theta)$ obtained from the neural network model after training for 1000 epochs, and predicted first derivative $u_x(x;\tilde{\theta)}$ obtained from the neural network model at the perturbed state, (b) absolute point-wise difference between $u_x(x;\theta)$ and $u_x(x;\tilde{\theta})$}
    \label{fig:model_prediction_first_derivative}
\end{figure*}

Furthermore, in Fig.~~\ref{fig:model_prediction_first_derivative}(a), we observe deviation of the first-order derivative of the prediction of our model before and after noise injection. This deviation as can be seen from Fig.~~\ref{fig:model_prediction_first_derivative}(b) is one order of magnitude higher than the one in Fig.~~\ref{fig:model_predictions}(b), which clearly shows that a first-order differential operator on a noisy output produces a much higher level of noise in the first order derivative.  Therefore, we deduce that noisy outputs result in even noisier derivatives, which in turn amplifies the noise in the physics loss term.
    
\begin{figure*}[!ht]
\centering
    \subfloat[]{\includegraphics[scale=0.650]{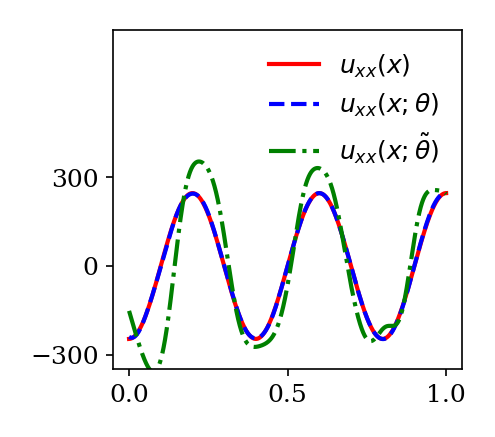}}\hspace{6em}
    \subfloat[]{\includegraphics[scale=0.650]{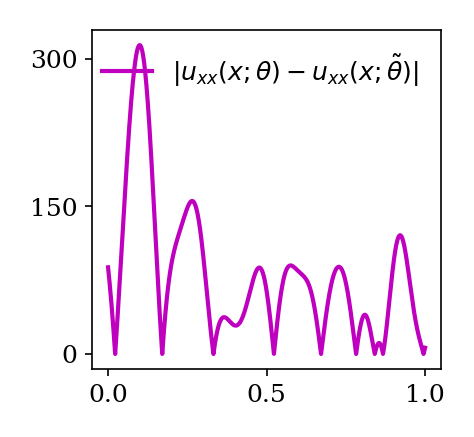}}\\
    \caption{Second-order derivative of the prediction obtained from the neural network model at different states: (a) comparison of the exact second derivative $u_x(x)$, predicted second derivative $u_x(x;\theta)$ obtained from the neural network model after training for 1000 epochs, predicted second derivative $u_x(x;\tilde{\theta)}$ obtained from the neural network model at the perturbed state, (b) absolute point-wise difference between $u_{xx}(x;\theta)$ and $u_{xx}(x;\tilde{\theta})$}
    \label{fig:model_prediction_second_derivative}
\end{figure*}

From Figs.~\ref{fig:model_prediction_second_derivative}(a) and (b), we observe that deviations become severe between the second-order derivative of the predictions obtained from our model before and after the noise injection. This illustrations reinforces our earlier conclusion that noisy outputs amplify noise in the derivatives such that the higher the order of the derivative operators, the larger the level of noise.
  
\begin{figure*}[!ht]
\centering
     \subfloat[]{\includegraphics[scale=0.650]{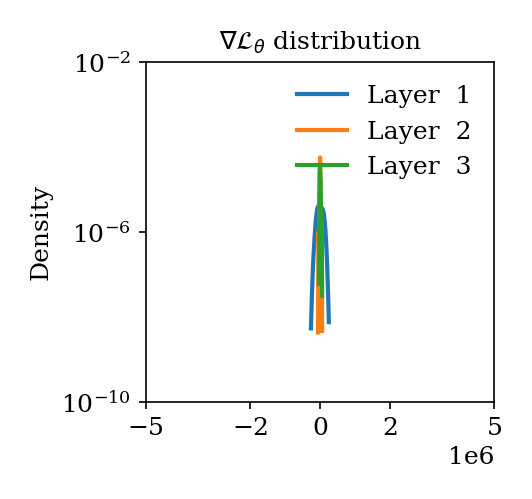}}\hspace{4em}
      \subfloat[]{\includegraphics[scale=0.650]{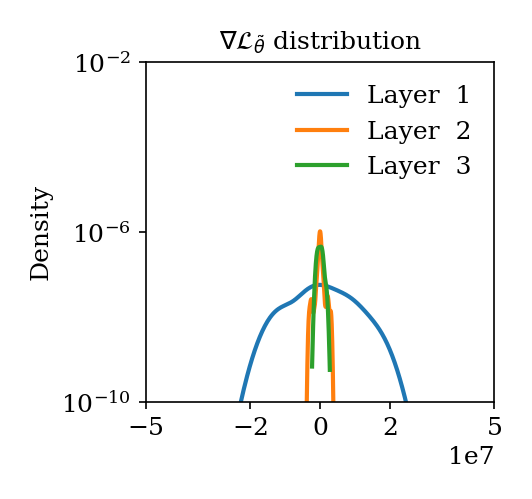}}
        \caption{Impact of a differential operator in amplification of noise for the calculation of back propagated gradients: (a) distribution of the gradients of the parameters of our neural network model after training for 1000 epochs, (b) distribution of the gradients of the parameters of our neural network model at the perturbed state.}
    \label{fig:model_param_grads}
\end{figure*}

Finally, from Fig.~~\ref{fig:model_param_grads}(a), we observe that the gradients of the parameters of the neural network before the noise injection are concentrated near zero, which shows that the optimizer has approached a local minimum. However, after the noise injection, we observe that the gradients of the parameters of the neural network have increased by an order of magnitude, which demonstrates the impact of physics loss on amplifying the perturbations and, consequently, corrupting the gradients. In other words, corrupting the gradients may result in impeding or preventing the optimizer from convergence since it will affect the parameters which in turn perturb the solution of the model. Therefore, we conclude that learning solutions arising from low-order derivative operators are favorably easier to learn than learning solutions arising from high-order derivative operators.

Next, we mathematically quantify the challenge of learning the solution of a well-posed PDE with neural networks. Consider the following functional relation
\begin{equation}
    F(x,u) = 0,
    \label{eq:functional_relation}
\end{equation}
where $x$ (input) is the set of data on which the solution $u$ (output) depends and $F$ is the functional relation between $x$ and $u$.  Following the presentation in \citet{quarteroni2010numerical}, a relative condition number can be defined as follows 
\begin{definition}[Relative condition number]

\begin{equation}
    \kappa(x) = \frac{\|\delta u \|/\|u\|}{\|\delta x \|/\|x\|}, \qquad \delta x  \neq 0,
    \label{eq:condition_number_formulat}
\end{equation}
\end{definition}
\noindent where $\kappa$ is the condition number, $\delta x$ is a small perturbation in the data, and $\delta u$ is the corresponding perturbation in the solution. If Eq.~\eqref{eq:functional_relation} admits a unique solution, then a mapping function $G$ between the set of data $x$ and the solution $u$ necessarily exists
\begin{equation}
    u = G(x),\quad F(G(x),x) = 0.
    \label{eq:resolvent}
\end{equation}
Hence, we can write the relative condition number as given in Eq.~\eqref{eq:condition_number_formulat} in the following form
\begin{equation}
   \kappa(x)  = \|G'(x)\|\frac{\|x\|}{\|G(x)\|},
   \label{eq:condition_formula}
\end{equation}
where $\|\cdot\|$ is a vector norm and $G'$ is the derivative of $G$ with respect to $x$. We use Eq.~\eqref{eq:condition_formula} to demonstrate  ill-conditioning of solutions to PDEs that may become challenging to learn with neural networks. 
\section{Proposed Method}\label{sec:Proposed_Method}
Consider the following second-order PDE
\begin{align}
-\text{div}( A \nabla u) + X u &= f, \quad \text{in} \quad \Omega \in \mathbf{R}^d,
\label{eq:original_pde}
\end{align}
subject to the following boundary conditions
\begin{align}
u = g_{D},~ \text{on}~ \Gamma_D, ~ \text{and}~ -\boldsymbol{n} \cdot A\nabla u = g_N, ~\text{on}~ \Gamma_N,
\label{eq:pde_boundary_conditions}
\end{align}
where $\Omega $ is the domain with its boundary $\partial \Omega = \Gamma_{D} \bigcup \Gamma_N$ with $\Gamma_D \bigcap \Gamma_N = \varnothing $.  $A(x)$ is a $d\times d$ symmetric matrix-valued function, $X$ is a differential operator of order at most one, and $\boldsymbol{n}$ is the outward normal unit vector to the boundary. $f$ , $g_D$ and $g_N$ are source functions in $\Omega$, $\Gamma_D$ and $\Gamma_N$, respectively. 

The problem defined by Eqs.~\eqref{eq:original_pde} and \eqref{eq:pde_boundary_conditions} is generally non-symmetric and does not have any underlying minimization principle. Following  the work of \citet{cai2020deep}, we introduce a flux parameter $\boldsymbol{\sigma}$, the problem defined by \eqref{eq:original_pde} and \eqref{eq:pde_boundary_conditions} can be written as a first-order system of PDE as follows: 
\begin{equation}
    \mathcal{D} = -\text{div}(\boldsymbol{\sigma}) + X u - f 
    \label{eq:first_order_pde}
\end{equation}
subject to the the following constraints 
\begin{align}
    \mathcal{F} &= \boldsymbol{\sigma} - A \nabla u, \hspace{1.5em} \text{in} \quad \Omega,\label{eq:original_flux_constraint}\\
    \mathcal{B} &= u - g_{D}, \hspace{2.5em} \text{on} \quad \Gamma_D,\label{eq:original_dirichlet_constraints} \\
    \mathcal{N} &=\boldsymbol{n} \cdot \boldsymbol{\sigma} + g_N,  \hspace{1em} \text{on} \quad \Gamma_N.
    \label{eq:original_neumann_constraints}
\end{align}
By introducing the flux variable $\boldsymbol{\sigma}$, we reduced the second-order PDE in Eq. \eqref{eq:original_pde} to the first-order PDE in Eq. \eqref{eq:first_order_pde}, which can then be used to learn by a single neural network model under the constraints presented in Eqs. \eqref{eq:original_flux_constraint}-\eqref{eq:original_neumann_constraints}.  Consider the following constrained optimization problem

\begin{equation}
    \min_{\theta} \sum_{i=1}^{N_{\mathcal{D}}}\|\mathcal{D}(x^{(i)};\theta)\|_2^2
\end{equation}
subject to the following constraints 

\begin{align}
    -\epsilon \le \mathcal{B}(x^{(i)};\theta) \le \epsilon, \quad \forall i = 1, \cdots, N_{\mathcal{B}},\\
    -\epsilon \le \mathcal{N}(x^{(i)};\theta) \le \epsilon, \quad \forall i = 1, \cdots, N_{\mathcal{N}},\\
    -\epsilon \le \mathcal{F}(x^{(i)};\theta) \le \epsilon, \quad \forall i = 1, \cdots, N_{\mathcal{F}},
\end{align}
where $\epsilon > 0 $ is a small positive number. Our goal in the above constrained optimization is to minimize an objective function such that constraints are satisfied within a small range of errors $(-\epsilon, \epsilon)$. Without loss of generality, we can simplify our constraints as follows 
\begin{align}
    \phi(\mathcal{B}(x^{(i)};\theta)) &\le \epsilon, \quad \forall i = 1, \cdots, N_{\mathcal{B}},\\
   \phi(\mathcal{N}(x^{(i)};\theta)) &\le \epsilon, \quad \forall i = 1, \cdots, N_{\mathcal{N}},\\
   \phi(\mathcal{F}(x^{(i)};\theta)) &\le \epsilon, \quad \forall i = 1, \cdots, N_{\mathcal{F}},
\end{align}
where $\phi \in [0,\infty)$ is a convex distance function (i.e. absolute value function or quadratic function). Since the minimum of our distance function occurs at the minimum of its inputs, the above formulations for our constraints are equivalent. Therefore, by pushing the $\epsilon \rightarrow 0 $, we get equality constraints as follows
\begin{align}
    \phi(\mathcal{B}(x^{(i)};\theta)) &= 0, \quad \forall i = 1, \cdots, N_{\mathcal{B}},\\
   \phi(\mathcal{N}(x^{(i)};\theta)) &= 0, \quad \forall i = 1, \cdots, N_{\mathcal{N}},\\
   \phi(\mathcal{F}(x^{(i)};\theta))&= 0, \quad \forall i = 1, \cdots, N_{\mathcal{F}},
\end{align}

Next, we employ the augmented Lagrangian method to formulate a dual unconstrained optimization problem suitable for training neural networks. Given a set of $N_{\mathcal{D}}$ residual points $\{x^{(i)} \}_{i=1}^{N_{\mathcal{D}}}$ in the domain $\Omega$, $N_{\mathcal{F}}$ residual points $\{x^{(i)} \}_{i=1}^{N_{\mathcal{F}}}$ in the domain $\Omega$,  $N_{\mathcal{B}}$ boundary points $\{(x^{(i)},g_D^{(i)}) \}_{i=1}^{N_{\mathcal{B}}}$ in $\Gamma_D$ and $N_{\mathcal{N}}$ boundary points $\{(x^{(i)},g_N^{(i)}) \}_{i=1}^{N_{\mathcal{N}}}$ in $\Gamma_N$, we can write the following objective function 
\begin{equation}
    \mathcal{L}_{\mu}(\theta;\lambda) = \mathcal{L}_{\mathcal{D}}(\theta) + \mathcal{L}_{\mathcal{F}}(\theta;\lambda_{\mathcal{F}}) + \mathcal{L}_{\mathcal{B}}(\theta;\lambda_{\mathcal{B}})  + \mathcal{L}_{\mathcal{N}}(\theta;\lambda_{\mathcal{N}}) + \pi(\theta),
\end{equation}
where 
\begin{align}
    \mathcal{L}_{\mathcal{D}}(\theta)  &= \sum_{i=1}^{N_{\mathcal{D}}}\|\mathcal{D}(x^{(i)};\theta)\|_2^2,\label{eq:pde_loss}\\
    \mathcal{L}_{\mathcal{F}}(\theta;\lambda_{\mathcal{F}}) &= \sum_{i=1}^{N_{\mathcal{F}}}\lambda_{\mathcal{F}}^{(i)} \phi(\mathcal{F}(x^{(i)};\theta)),\label{eq:flux_loss}\\
    \mathcal{L}_{\mathcal{B}}(\theta;\lambda_{\mathcal{B}}) &= \sum_{i=1}^{N_{\mathcal{B}}}\lambda_{\mathcal{B}}^{(i)} \phi(\mathcal{B}(x^{(i)};\theta)),\label{eq:dirichlete_boundary_loss}\\
    \mathcal{L}_{\mathcal{N}}(\theta;\lambda_{\mathcal{N}}) &= \sum_{i=1}^{N_{\mathcal{N}}}\lambda_{\mathcal{N}}^{(i)} \phi(\mathcal{N}(x^{(i)};\theta)),\label{eq:neumann_boundary_loss}
\end{align}
and the penalty term is as follows 
\begin{align}
    \pi(\theta) &=  \frac{1}{2}\pi_{\mathcal{F}}(\theta) + \frac{\mu}{2}\pi_{\mathcal{B}}(\theta) + \frac{\mu}{2} \pi_{\mathcal{N}}(\theta),
    \label{eq:penalty_function}\\
    \pi_{\mathcal{F}}(\theta) &=\sum_{i=1}^{N_{\mathcal{F}}}\| \phi(\mathcal{F}(x^{(i)};\theta))\|^2,\\
    \pi_{\mathcal{B}}(\theta) &=\sum_{i=1}^{N_{\mathcal{B}}}\| \phi(\mathcal{B}(x^{(i)};\theta))\|^2,\\
    \pi_{\mathcal{N}}(\theta) &=\sum_{i=1}^{N_{\mathcal{N}}}\| \phi(\mathcal{N}(x^{(i)};\theta))\|^2,
\end{align}
where $\mu$ is a positive penalty parameter, $\lambda_{\mathcal{F}}$, $\lambda_{\mathcal{D}}$ and $\lambda_{\mathcal{N}}$ are vectors of Lagrange multipliers to enforce Eqs. \eqref{eq:original_flux_constraint}, \eqref{eq:original_dirichlet_constraints} and \eqref{eq:original_neumann_constraints}, respectively. These multipliers are responsible for adjusting the global learning rate based on their corresponding loss functions. These vectors of Lagrange multipliers may also be viewed as preconditioning matrices that are diagonal with their elements set to the corresponding vectors of Lagrange multipliers. $\phi$ is a convex distance function. Due to the quadratic penalty function as in Eq.\eqref{eq:penalty_function}, we employ Huber function \cite{huber1992robust} as our distance function for its robustness to outliers in comparison with a quadratic distance function as used in our previous work \cite{PECANN_2022}. Using the update rule for the augmented Lagrangian method, we update the Lagrange multipliers as follows
\begin{align}
  \lambda_{\mathcal{F}}^{(i)} &\leftarrow \lambda_{\mathcal{F}}^{(i)} +  \phi(\mathcal{F}(x^{(i)};\theta)) , \forall i = 1,\cdots, N_{\mathcal{F}},\label{eq:lambda_f_rule}
  \\
\lambda_{\mathcal{B}}^{(i)} &\leftarrow \lambda_{\mathcal{B}}^{(i)} + \mu \phi(\mathcal{B}(x^{(i)};\theta)) , \forall i = 1,\cdots, N_{\mathcal{B}},\\
\lambda_{\mathcal{N}}^{(i)} &\leftarrow \lambda_{\mathcal{N}}^{(i)} + \mu \phi(\mathcal{N}(x^{(i)};\theta)) , \forall i = 1,\cdots, N_{\mathcal{N}},
\end{align}
where $\xleftarrow{}$ indicates an optimization step. We also provide a simple training algorithm as follows
\IncMargin{1em}
\begin{algorithm}
\SetAlgoLined
\SetKw{KwInput}{Input:}
\SetKw{KwOutput}{Output:}
\KwInput{$\theta^0, \mu_{max}$}\\
$\lambda_{\mathcal{B}}, \lambda_{\mathcal{I}}, \lambda_{\mathcal{F}}  \leftarrow 0 $ \\
$\mu \leftarrow 1$\\
\KwOutput{$\theta^*$}\\
\BlankLine
    $\theta^* \leftarrow \underset{\theta}{\mathrm{argmin}}~ \mathcal{L}_{\mu}(\theta;\lambda_{\mathcal{F}},\lambda_{\mathcal{B}},\lambda_{\mathcal{I}})$\\
    $\mu \leftarrow \min(2\mu, \mu_{max})$\\
    $\lambda_{\mathcal{B}} \leftarrow \lambda_{\mathcal{B}} + \mu \phi(\mathcal{B}(\boldsymbol{x};\theta))$\\
    $\lambda_{\mathcal{I}} \leftarrow \lambda_{\mathcal{I}} + \mu \phi(\mathcal{I}(\boldsymbol{x};\theta))$\\
    $\lambda_{\mathcal{F}} \leftarrow \lambda_{\mathcal{F}} +  \phi(\mathcal{F}(\boldsymbol{x};\theta))$
 \caption{Training algorithm}
 \label{alg:TrainingAlgorithm}
\end{algorithm}

\subsection{Performance Metrics}
We adopt the following metrics for evaluating the prediction of our models. Given an $n$-dimensional vector of predictions $\boldsymbol{\hat{u}} \in \mathbf{R}^n$ and an $n$-dimensional vector of exact values $\boldsymbol{u} \in \mathbf{R}^n$, we define a relative Euclidian or $L^2$ norm
\begin{align}
     \epsilon_r(\hat{u},u) &= \frac{\|\hat{\boldsymbol{u}} - \boldsymbol{u}\|_2}{\|\boldsymbol{u}\|_2}, \\
    \label{eq:relativeL2Error}
    \epsilon_{\infty}(\hat{u},u) &= \| \boldsymbol{\hat{u}} - \boldsymbol{u}\|_{\infty},\\
    \text{MAE} &= \frac{1}{n}\sum_{i=1}^{n}(\hat{\boldsymbol{u}}^{(i)} - \boldsymbol{u}^{(i)})^2
\end{align}
where $\|\cdot \|_2$ denotes the Euclidean norm and $\| \cdot \|_{\infty}$ denotes the maximum norm.

\section{Numerical Experiments}\label{sec:Numerical_Experiments}
We apply our proposed method to learn the solution of several benchmark PDEs that are prevalent in computational physics. We also compare our results with other published results to highlight the marked improvements in accuracy levels due to our proposed method. 

\subsection{Heat Transfer in Composite Materials}
In this section, we study a PDE problem with a non-smooth solution. Because the solution is non-smooth, we cannot directly apply the strong for of the PDE. However, we can relax the smoothness requirement by introducing an auxliary flux parameter to obtain a system of first-order PDE. We study a typical heat transfer in a composite material where temperature and heat fluxes are matched across the interface \cite{baker1985heat}. Consider a one-dimensional heat equation, which is also considered in the work osf \citet{cai2020deep},
\begin{align}
    - \frac{\partial }{\partial x}[a(x) \frac{\partial u(x)}{\partial x}] &= f(x), ~x \in \Omega = (0,1),
    \label{eq:heat_pde}
    \\
    u(x) &= 0, ~ x \in \partial \Omega = \{0,1\},
\end{align}
where $a=1$ for $x \in (0,\frac{1}{2})$ and $a=k$ for $x \in (\frac{1}{2},1)$.
\begin{equation}
f(x) = 
\begin{cases} 
      8k(3x-1), & x \in (0,\frac{1}{2}) \\
      2k(k+1), & x \in (\frac{1}{2},1) 
   \end{cases}
\end{equation}

\begin{equation}
u(x) = 
\begin{cases} 
      4k x^2(1-x), & x \in (0,\frac{1}{2}) \\
      [2(k+1)x-1](1-x), & x \in (\frac{1}{2},1).
   \end{cases}
   \label{eq:exact_heat_solution}
\end{equation}
Since the solution given in Eq.~\eqref{eq:exact_heat_solution} is non-smooth, we cannot directly use Eq.~\eqref{eq:heat_pde}. We can reduce Eq.~\eqref{eq:heat_pde} into a system of first-order differential equations by introducing an auxiliary flux parameter $\sigma = -a u_x$. We use a fully connected neural network architecture, which consists of a one hidden layer with 32 neurons and sigmoid activation functions. Similar to \cite{cai2020deep}, we sample $N_{\Omega} = 500$ residual points from the interior part of the domain only once. Our distance function in this problem is the Huber function \cite{huber1992robust} and we adopt the L-BFGS  optimizer \cite{nocedal1980updating} with its default parameters and \emph{strong Wolfe} line search function that are built in PyTorch framework \cite{paszke2019pytorch}. We train our network for 20000 epochs with our safeguarding penalty parameter $\mu_{\max} = 10^4$. 

\begin{figure}[!h]
\centering
  \includegraphics[scale=0.650]{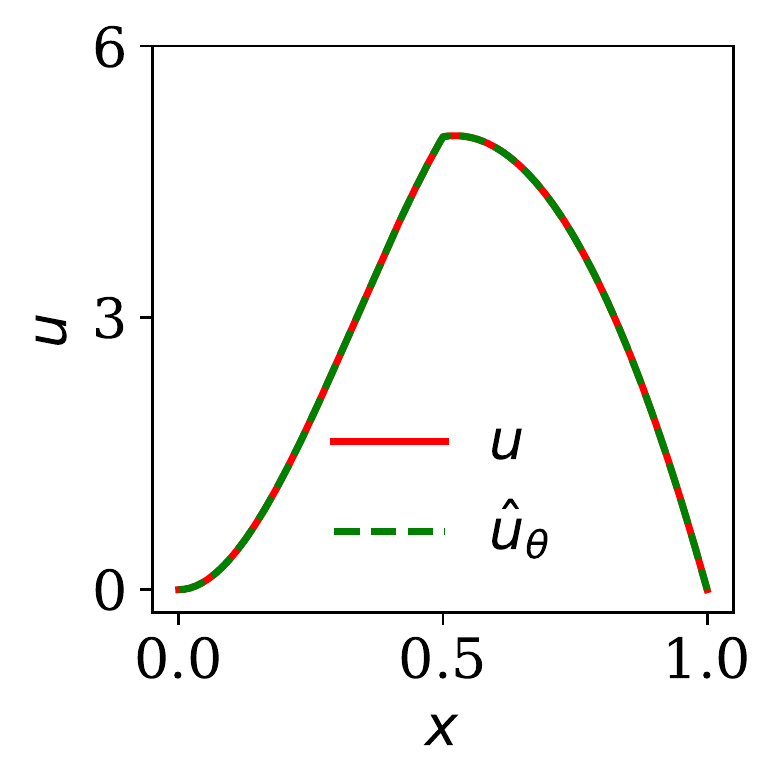}
       \caption{Heat transfer in composite medium: exact solution $u$ vs. predicted solution $\hat{u}_{\theta}$}
    \label{fig:heat_solution}
\end{figure}

\begin{figure}[!h]
\centering
\includegraphics[scale=0.650]{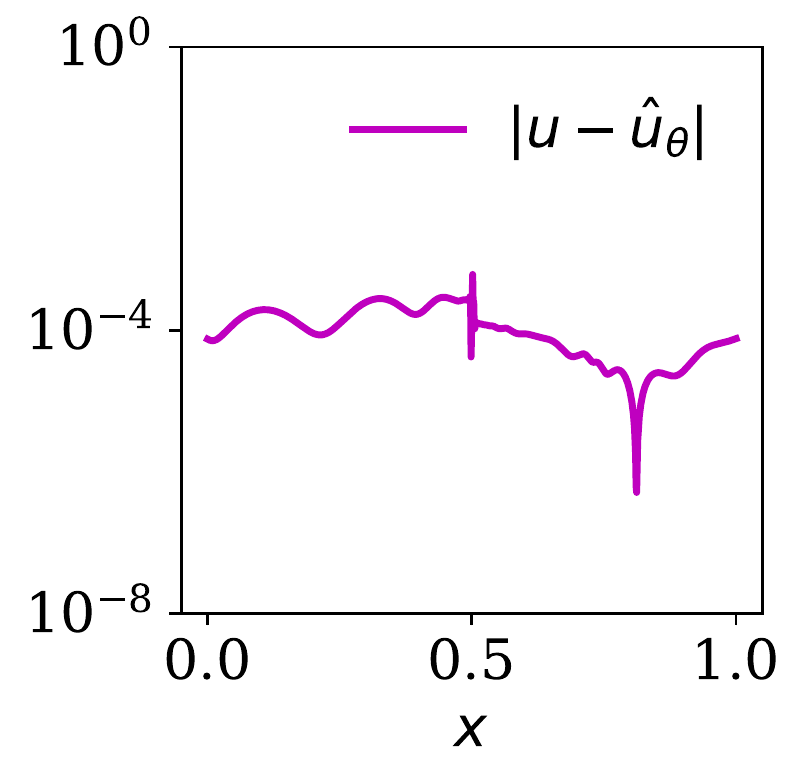}
       \caption{Heat transfer in composite medium: absolute point-wise error of predicted solution $\hat{u}_{\theta}$.}
    \label{fig:heat_error}
\end{figure}

We present the prediction of our neural network model in Fig.~\ref{fig:heat_solution}. 
From Fig.~\ref{fig:heat_error}, we observe that our neural network model has successfully learned the underlying solution. 

\begin{figure}[!ht]
\centering
    \includegraphics[scale=0.65]{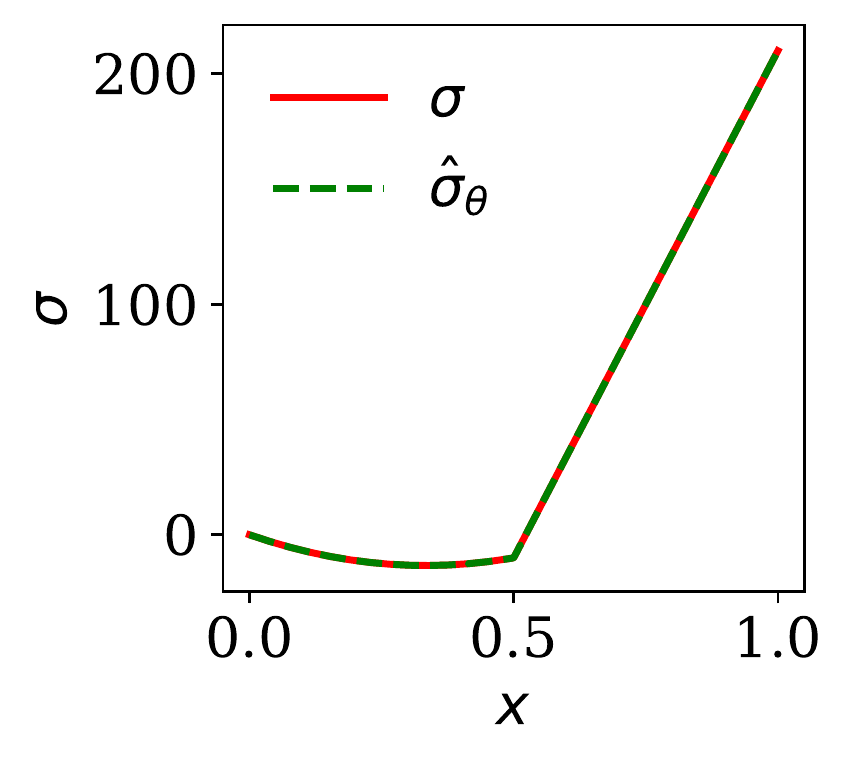}
    \caption{Heat transfer in composite medium: exact flux distribution $\sigma$ vs. predicted flux distribution $\hat{\sigma}_{\theta}$}
    \label{fig:heat_flux}
\end{figure}

\begin{figure}[!ht]
\centering
    \includegraphics[scale=0.65]{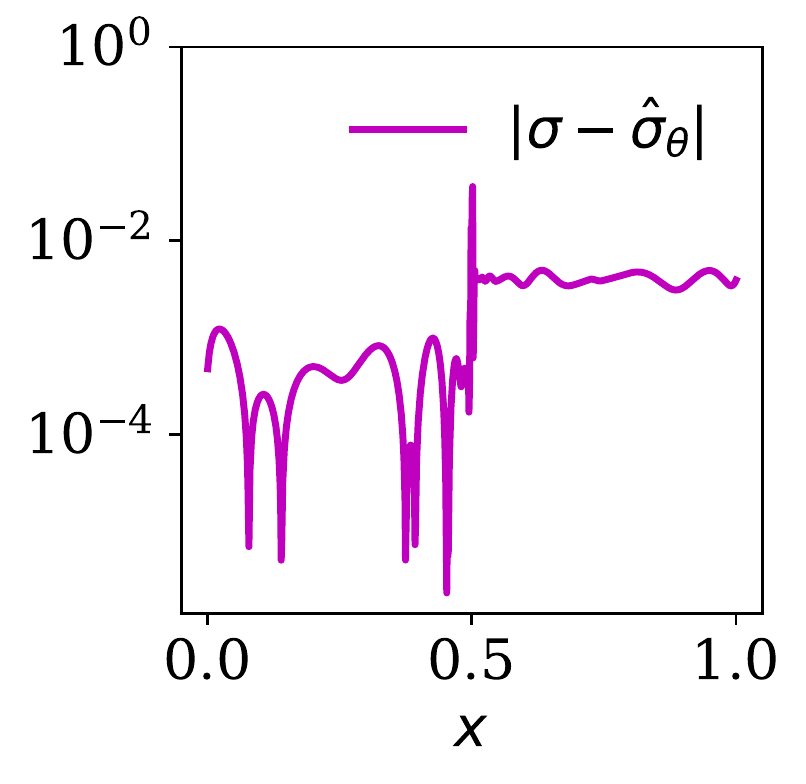}
    \caption{Heat transfer in composite medium absolute point-wise error distribution of predicted flux.}
    \label{fig:heat_flux_error}
\end{figure}

\begin{figure}[!ht]
\centering
    \includegraphics[scale=0.70]{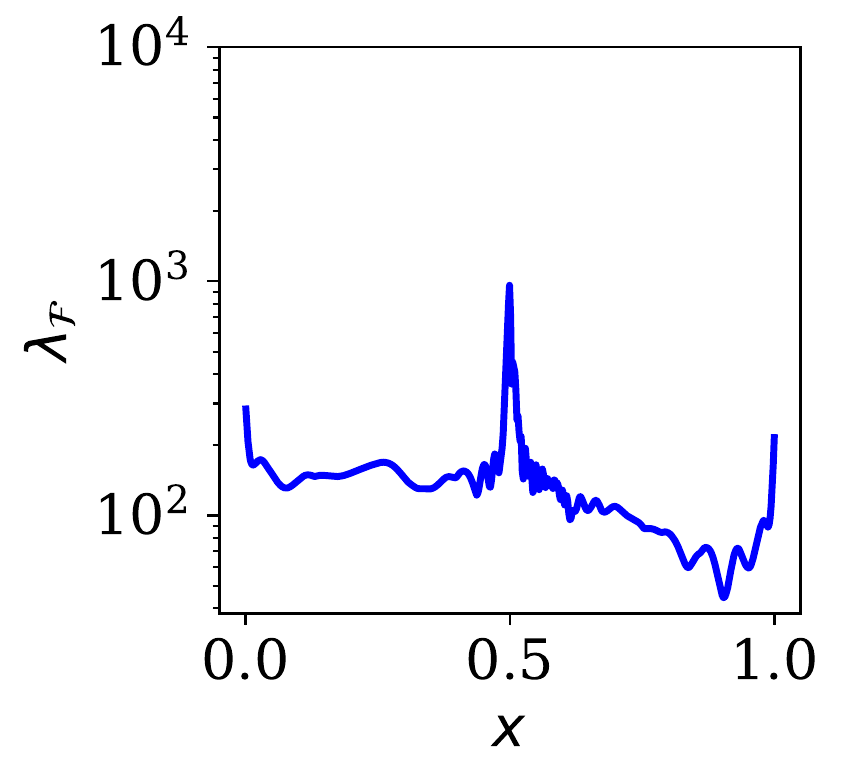}
    \caption{Heat transfer in composite medium distribution of Lagrange multipliers for local flux constraints.}
    \label{fig:interface_lagrange_multipliers}
\end{figure}

Next, we present the predicted flux obtained from our neural network model in Fig.~\ref{fig:heat_flux}. From Fig.~\ref{fig:heat_flux_error}, we observe that our neural network model has successfully learned the underlying flux. Finally, we present the distribution of Lagrange multipliers for enforcing the flux constraints in Fig.~\ref{fig:interface_lagrange_multipliers}. 

From Fig.~\ref{fig:interface_lagrange_multipliers}, we observe a spike in the distribution of Lagrange multipliers $\lambda_{\mathcal{F}}$ at the interface where fluxes are matched. This shows that our model has adaptively learned to focus on regions where the fluxes are challenging to learn to ensure the uniform feasibility of flux constraint across the domain. 

We also present a summary of the medians of error norms over three independent trials with Xavier initialization scheme \cite{glorot2010understanding} in Table~\ref{tb:interface}. The results indicate that our method achieves two orders of magnitude lower norms of error than the method presented in \cite{cai2020deep}. We also emphasize that in our approach we only use a single layer of the same neural network architecture used in \citet{cai2020deep}, whereas \citeauthor{cai2020deep} use a composite architecture by design. 

\begin{table*}[!ht]
\centering
\caption{Summary of the median of error norms for the interface problem.}
\label{tb:interface}
\vspace{2pt}
\resizebox{0.6\textwidth}{!}{%
\begin{tabular}{@{}lccccc@{}}
\toprule
\multicolumn{1}{c}{Models} &
  \multicolumn{1}{c}{$\epsilon_{r}(u,\hat{u})$}
  &
   \multicolumn{1}{c}{$\epsilon_{r}(\sigma,\hat{\sigma})$}&
  \multicolumn{1}{c}{No. Parameters}&
  \\ \midrule
Ref. \cite{cai2020deep} (sigmoid)  & $7.137 \times 10 ^{-3}$  & $1.870 \times 10 ^{-3}$ & $2962$\\
Current (sigmoid)  & $\boldsymbol{4.654 \times 10^{-5}}$  & $\boldsymbol{4.457 \times 10 ^{-5}}$ & $\boldsymbol{130}$\\
\bottomrule
\end{tabular}}
\end{table*}

\subsection{Convection-dominated convection–diffusion equation}
A wide range of physical problems involves diffusive and convective (transport) processes. Standard numerical methods work well when diffusion dominates convection. However, standard numerical methods such as finite differences or standard Galerkin finite elements become unstable when convection effects dominates over diffusion effects \cite{fiard1998first}.  Consider the following problem that is also studied in the work of \citet{van2020optimally}

\begin{align}
    v \frac{d u(x)}{d x} + \alpha \frac{d^2 u(x) }{dx^2} &= 0, \qquad \text{in} \quad \Omega,
    \label{eq:convection_diffusion_pde}\\
    u &= g(x), \quad \text{on} \quad \partial \Omega
    \label{eq:convection_diffusion_bc}
\end{align}
where $v = 1$, $\alpha$ is the diffusivity coefficient, $\Omega = \{x~|~ 0 \le x \le 1 \}$ with its boundary $\partial \Omega$. The analytical solution of the above equation is given as follows 
\begin{equation}
    u(x) = \frac{e^{-\frac{v x}{\alpha}}}{1 - e^{-\frac{v}{\alpha}}} - \frac{1}{2}.
    \label{eq:convection_diffusion_exact}
\end{equation}

\begin{figure}[!h]
\centering
    \includegraphics[scale=0.70]{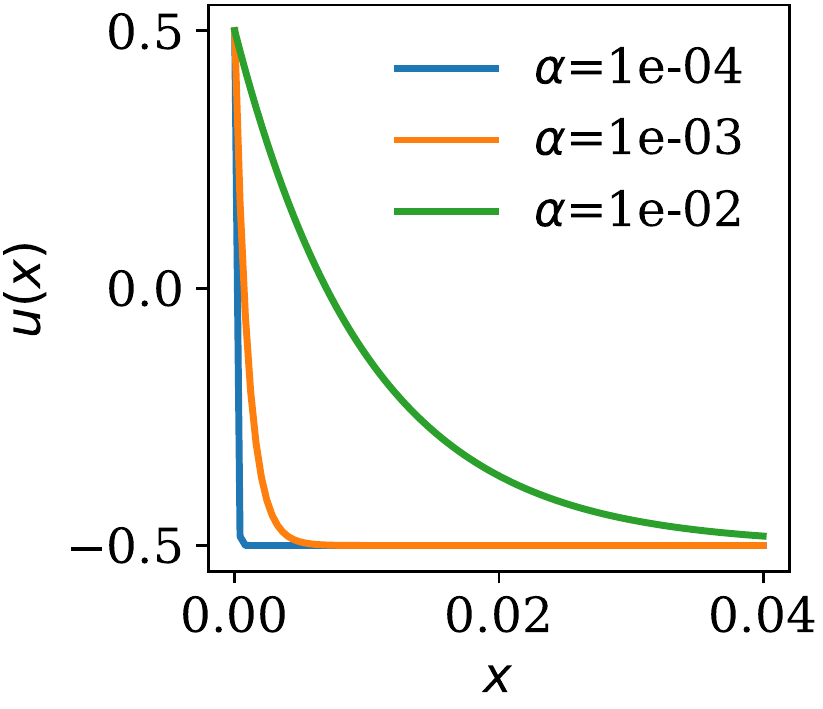}
    \caption{Convection dominated convection-diffusion equation: (a) exact solution for various $\alpha$}
    \label{fig:conv_diff_condition_vs_alpha}
\end{figure}

From Eq.~\eqref{eq:convection_diffusion_exact}, we observe that the size of the boundary layer is proportional to the diffusivity coefficient $\alpha$. Therefore, the solution becomes challenging as $\alpha$ is decreased. Solution of convention-dominated convection-diffusion problems can result in boundary layers in which the solution behaves drastically differently in a small part of the domain. Therefore, learning a solution for these types of PDEs can become challenging. From Fig.~\ref{fig:conv_diff_condition_vs_alpha}, we observe that as $\alpha$ decreases, the solution drastically changes, which becomes challenging to capture with conventional numerical methods. 

\begin{figure}[!h]
\centering
    \includegraphics[scale=0.70]{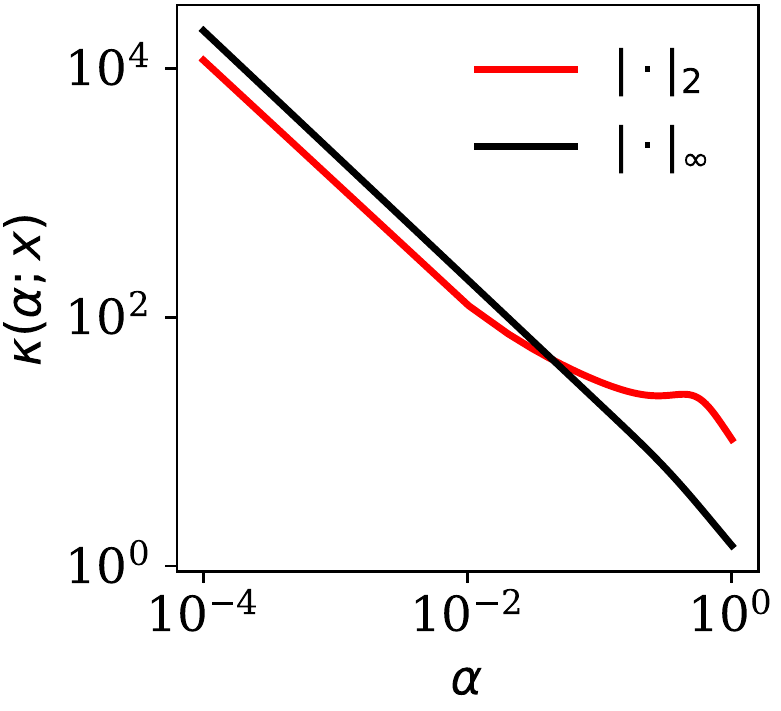}
    \caption{Convection dominated convection-diffusion equation: (a)  condition number }
    \label{fig:conv_diff_condition_condition_number}
\end{figure}

To mathematically quantify this challenge, we derive a condition number for this problem in terms of $\alpha$. To achieve this, we create a vector of inputs $x$ with 100 elements and evaluate the solution $G(x)$ from Eq.~\eqref{eq:convection_diffusion_exact} with its derivative $G'(x)$ exactly calculated. We then use Eq.~\eqref{eq:condition_formula} to approximate a condition number for decreasing values of $\epsilon$. The resulting condition number is presented in Fig.~\eqref{fig:conv_diff_condition_condition_number}. From Fig.~~\eqref{fig:conv_diff_condition_condition_number}, we observe that as we decrease $\alpha$, the condition number of the problem increases, which indicates the increasing complexity of learning the solution.

Having discussed the complexity of learning the solution of this problem, we use neural networks to learn the solution for of a challenging case for which $\alpha = 10^{-4}$. We achieve this by introducing an auxiliary flux parameter $\sigma(x) = -\alpha \frac{du(x)}{dx}$ to reduce Eq.~\eqref{eq:convection_diffusion_pde} to a system of first-order partial differential equations. For this problem, we use a fully connected feed-forward neural network with four hidden layer and 20 neurons as used in \cite{van2020optimally}.  Our network employs tangent hyperbolic non-linearity and has one input and two outputs corresponding to $u$ and $\sigma$. We adopt the L-BFGS optimizer with its default parameters that are built-in PyTorch framework \cite{paszke2019pytorch} and we train our network for 2000 epochs. We generate 2048 number of collocation points in the domain with its boundary data only once before training. We use Huber \cite{huber1992robust} function as our distance function and set our safeguarding penalty parameter $\mu_{\max}=10^4$. 

\begin{figure}[!h]
\centering
    \includegraphics[scale=0.70]{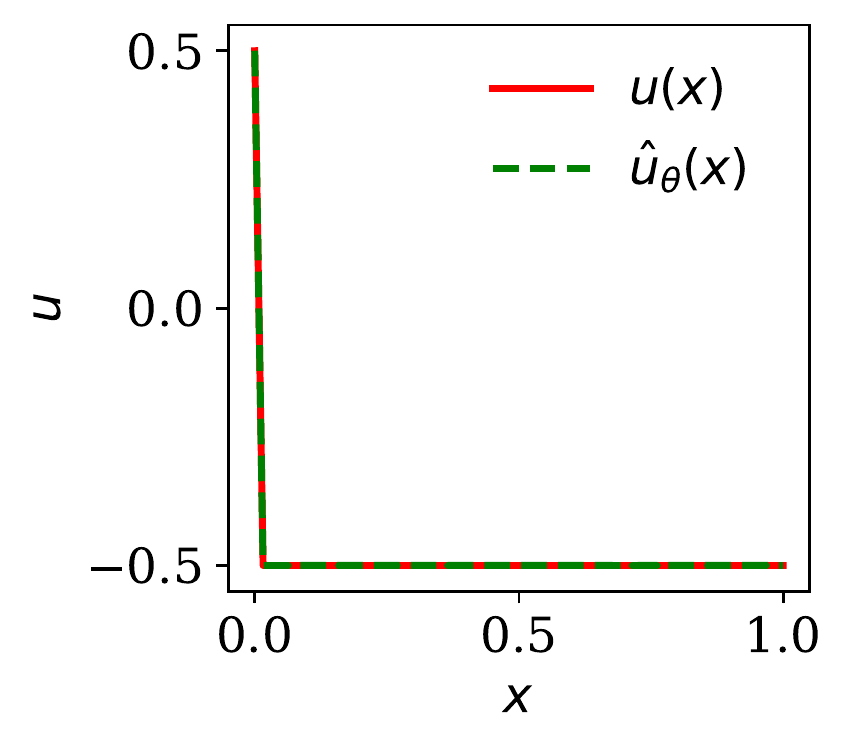}
    \caption{Convection-dominated convection diffusion equation: exact solution $u$ vs the predicted solution $\hat{u}_{\theta}$}
    \label{fig:convection_diffusion_solution}
\end{figure}

\begin{figure}[!h]
\centering
    \includegraphics[scale=0.70]{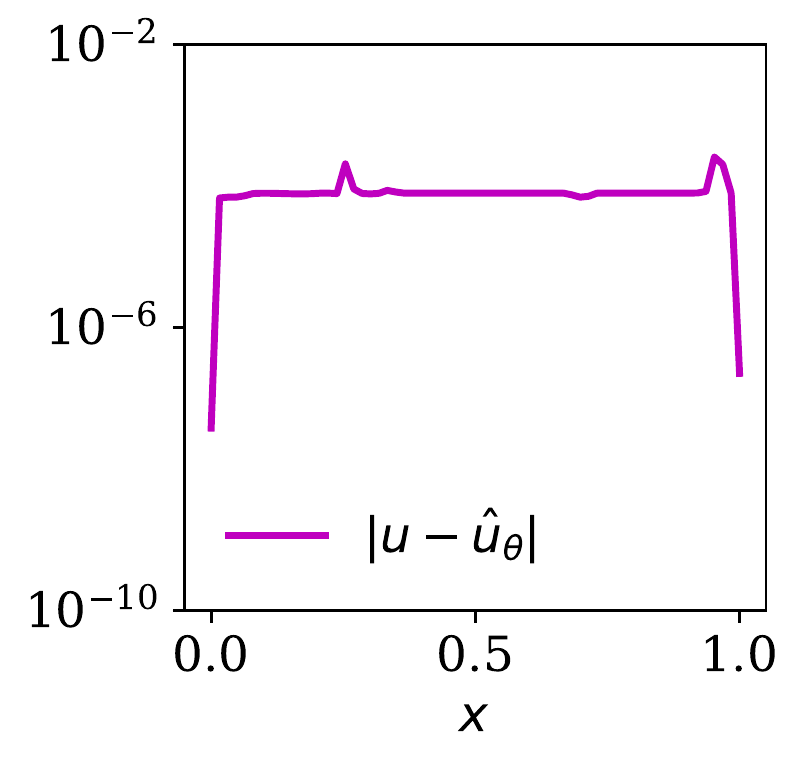} 
    \caption{Convection-dominated convection diffusion equation: absolute point-wise error of the predicted solution}
    \label{fig:convection_diffusion_error}
\end{figure}

\begin{figure}[!h]
\centering
    \includegraphics[scale=0.70]{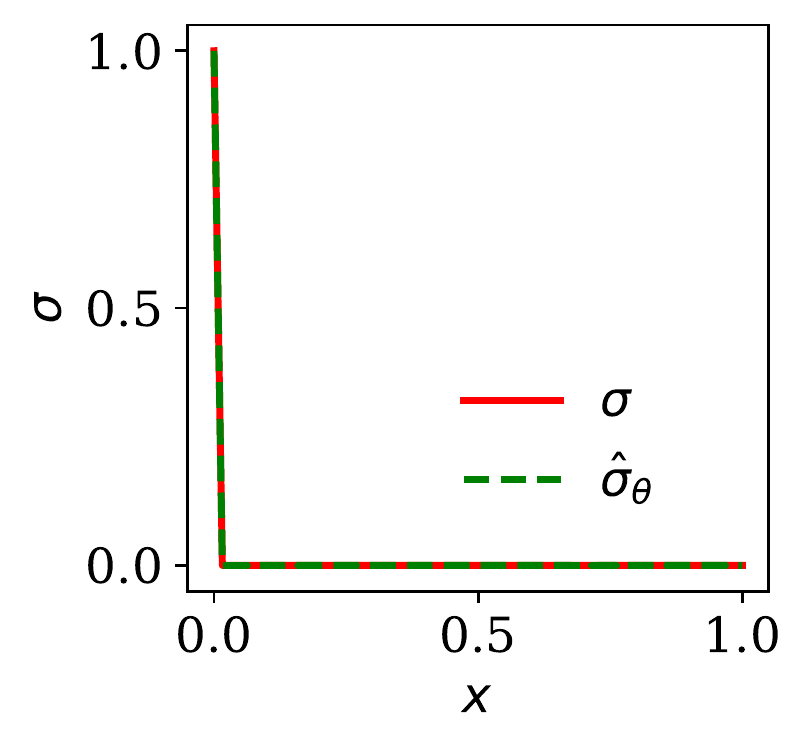}
    \caption{Convection-dominated convection diffusion equation: exact flux distribution $\sigma$ and the predicted flux distribution $\hat{\sigma}_{\theta}$}
    \label{fig:convection_diffusion_flux}
\end{figure}

\begin{figure}[!h]
\centering
    \includegraphics[scale=0.70]{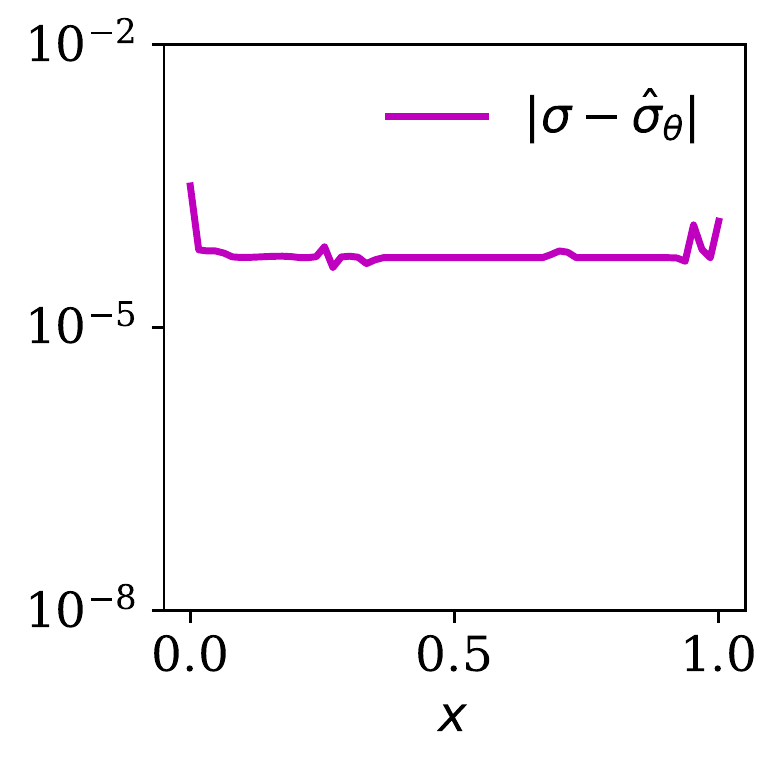}
    \caption{Convection-dominated convection diffusion equation: absolute point-wise error distribution of the predicted flux}
    \label{fig:convection_diffusion_flux_error}
\end{figure}

We present the prediction of our neural network model in Fig.~~\ref{fig:convection_diffusion_solution}. From Fig.~\ref{fig:convection_diffusion_error}, we observe that our model has accurately learned the underlying solution. Next, We present the prediction of our neural network model in Fig.~~\ref{fig:convection_diffusion_flux}.
From Fig.~\ref{fig:convection_diffusion_flux_error}, we observe that our model has accurately learned the underlying flux.

\begin{figure}[!ht]
\centering
    \includegraphics[scale=0.65]{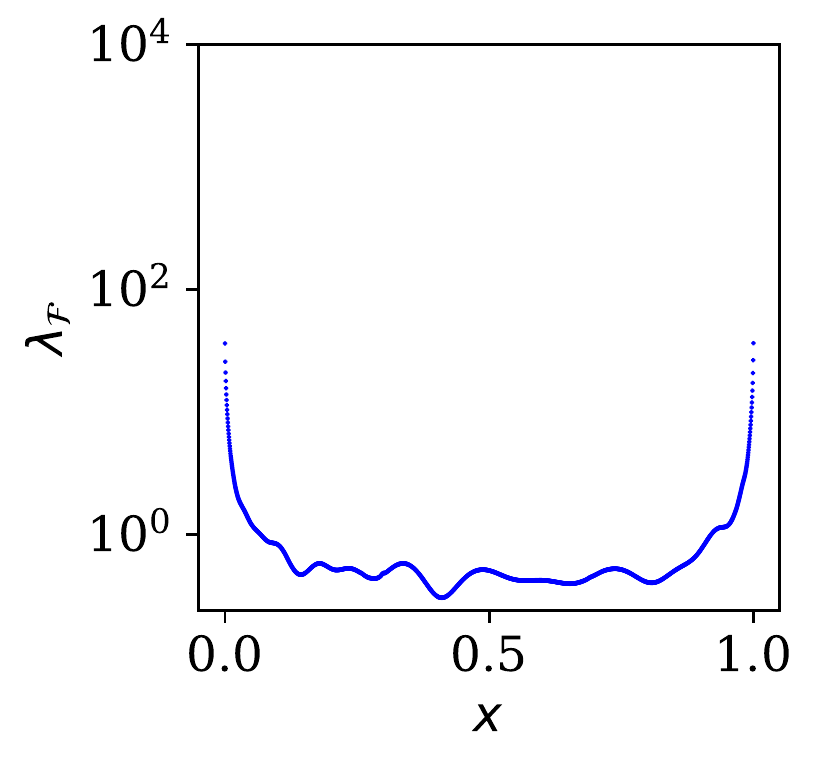}
    \caption{Convection-dominated convection diffusion equation: distribution of Lagrange multipliers for local flux constraints}
    \label{fig:convection_diffusion_lagrange_multiplier}
\end{figure}

Finally, we present the distribution of Lagrange multiplier in Fig.~\ref{fig:convection_diffusion_lagrange_multiplier}, which shows that our model has learned to focus on the regions that are challenging to learn.  

In addition, in Table~\ref{tb:convection_diffusion} we present the error norms obtained from different methods with random Xavier initialization scheme \cite{glorot2010understanding} . We observe that the relative error $\epsilon_r(u,\hat{u})$ obtained from our model is eighth orders of magnitude lower than the one obtained in \cite{van2020optimally}. We are also able to achieve this level of high accuracy with a single layer of their neural network architecture trained for one-tenth of the number of epochs used in \cite{van2020optimally}.

\begin{table*}[!ht]
\centering
\caption{Convection-diffusion: summary of error norms obtained from different methods}
\label{tb:convection_diffusion}
\vspace{2pt}
\resizebox{\textwidth}{!}{%
\begin{tabular}{@{}lcccccccc@{}}
\toprule
 \multicolumn{1}{c}{Method}&
 \multicolumn{1}{c}{$\alpha$}&
  \multicolumn{1}{c}{$\epsilon_{r}(u,\hat{u})$}&
  \multicolumn{1}{c}{$\epsilon_{\infty}(u,\hat{u})$}&
  \multicolumn{1}{c}{Sampling Strategy}&
  \multicolumn{1}{c}{Epochs}&
  \\ \midrule
  Optimal Loss Weight\cite{van2020optimally}& $10^{-4}$&$1.15 \times 10^{0} $&$2.00 \times 10^{0} $& Adaptive&$20\times 10^3$\\
 Magnitude Normalization.\cite{van2020optimally}& $10^{-4}$&$1.91 \times 10^{0} $ &$3.51 \times 10^{1}$& Adaptive  & $20\times 10^3$\\
 Proposed method & $\boldsymbol{10^{-4}}$& $\boldsymbol{1.78 \times 10^{-4}}$&$\boldsymbol{2.55 \times 10^{-4}}$& Uniform  & $\boldsymbol{2\times 10^3}$\\
\bottomrule
\end{tabular}}
\end{table*}

\subsection{Reaction Diffusion Equation}
Reaction-diffusion equations describe the behaviour of a wide range of chemical systems in which diffusion competes with the production of material by reaction. Other typical applications include systems where heat (or fluid) is produced and is diffused away from the source. In this section, we study a benchmark problem presented in \cite{krishnapriyan2021characterizing} and demonstrate two and three orders of magnitude improvement in comparison with the results obtained by \cite{krishnapriyan2021characterizing}. Consider the following partial differential equation
\begin{align}
        \frac{\partial u}{\partial t} -\nu \frac{\partial^2 u}{\partial x^2} - \rho u ( 1 -u)&= 0, ~\forall (x,t) \in \Omega \times [0,1],\\
        u(x,0) &= h(x), ~ \forall x \in \partial \Omega,\\
        u(0,t) &= u(2 \pi ,t) ~\forall t \in [0,1],\\
        u_x(0,t) &= u_x(2 \pi ,t) ~\forall t \in [0,1]
        \label{eq:reaction_diffusion_PDE}
\end{align}

where $\nu=6$, $\rho = 5$, $\Omega = \{ x ~ | ~ 0 \le x < 2\pi\}$ and $\partial \Omega$ is its boundary.
\begin{align}
    h(x) = e^{\frac{-(x - x_0)^2}{2\sigma^2}},
\end{align}
where $x_0 = \pi$ and $\sigma = \pi/4$.

\begin{align}
    &\mathcal{F}(x,t)= \frac{\partial u(x,t)}{\partial t}
    - \nu \frac{\partial^2 u(x,t)}{\partial x^2} - \rho u(x,t) ( 1 - u(x,t)), \\
    &\mathcal{B}(t) = u(0,t) - u(2\pi,t),\\
    &\mathcal{N}(t) = u_x(0,t) - u_x(2\pi,t),\\
    &\mathcal{I}(x) = u(x,0) - h(x),
\end{align}

\begin{figure}[!h]
\centering
\includegraphics[scale=0.60]{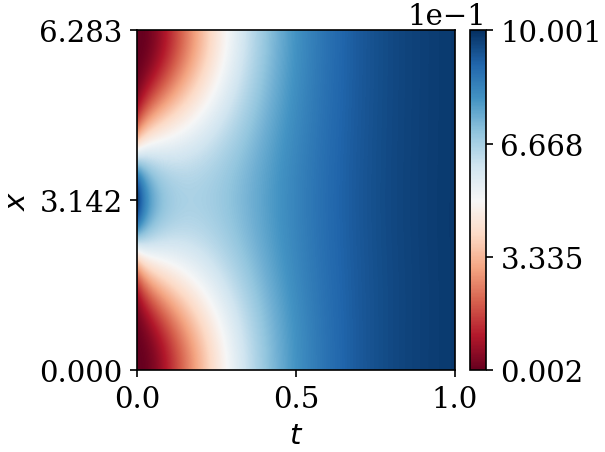}
\caption{Reaction diffusion equation: predicted solution}
\label{fig:reaction_diffusion_prediction}
\end{figure}

\begin{figure}[!h]
\centering
\includegraphics[scale=0.60]{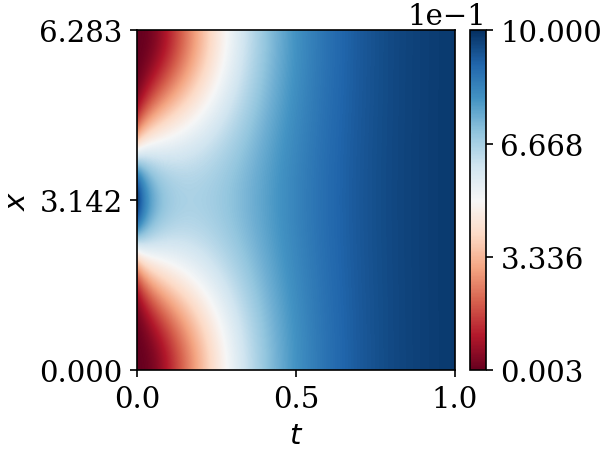}
\caption{Reaction diffusion equation: exact solution}
\label{fig:reaction_diffusion_exact}
\end{figure}

\begin{figure}[!h]
\centering
\includegraphics[scale=0.60]{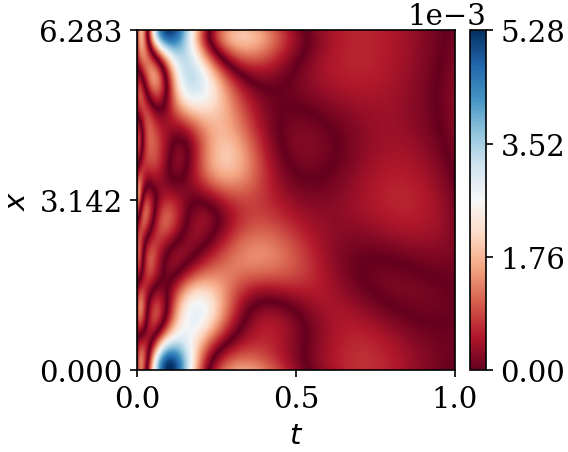}
\caption{Reaction diffusion equation:  absolute point-wise error $\epsilon_r(u,\hat{u}) = 9.326\times 10^{-4}$, MAE =  $6.312 \times 10^{-4}$}
\label{fig:reaction_diffusion_error}
\end{figure}

We use the same fully connected neural network architecture as in \cite{krishnapriyan2021characterizing}, which consists of four hidden layers with 50 neurons per layer and the tangent hyperbolic activation function. We use a Sobol sequence to sample $N_{\mathcal{F}} = 1024$ residual points from the interior part of the domain and $N_{\mathcal{B}} = 128$ from the boundaries and 
$N_{\mathcal{I}} = 128$ for approximating the loss on the initial condition
only once before training. Our optimizer is L-BFGS  \cite{nocedal1980updating} with its default parameters and \emph{strong wolfe} line search function that is built in PyTorch framework \cite{paszke2019pytorch}. We train our network for 500 epochs with our safeguarding penalty parameter $\mu_{\max} = 10^{4}$. 
We present the prediction obtained from our model in Fig.\ref{fig:reaction_diffusion_prediction}. We also present the exact solution in Fig.\ref{fig:reaction_diffusion_exact} for comparison purposes. From Fig.~\ref{fig:reaction_diffusion_error} we observe that our model has successfully learned the underlying solution. We should note that our approach is training the model only once for the entire state space unlike sequence-to-sequence training approach which trains the model at each time step $\Delta t$, which significantly increases the cost of training. 

We also present a summary of the error norms from our approach and state-of-the-art results presented in \cite{krishnapriyan2021characterizing} in Table~\ref{tb:reaction_diffusion}. We observe that results obtained by our method achieves a mean absolute error $MAE = 6.312 \times 10^{-4}$, which is two orders of magnitude lower than $1.28 \times 10^{-2}$ obtained by the method presented in \citet{krishnapriyan2021characterizing}. As seen from table~\ref{tb:reaction_diffusion}, sequence-to-sequence modeling does not produce better results with smaller time steps as expected. In addition, the computational complexity of the learning process is proportional to the number of time steps since the model is separately trained at each time step. In addition, errors made at an earlier time step corrupts the solution at a later time step since the solution of the earlier time step is used as an initial condition for training the model at the later time step. 

\begin{table*}[!h]
\centering
\caption{Reaction diffusion equation: summary of the relative $L_2$ and mean absolute error (MAE) obtained from different methods}
\label{tb:reaction_diffusion}
\vspace{2pt}
\resizebox{0.8\textwidth}{!}{%
\begin{tabular}{@{}lccccc@{}}
\toprule
\multicolumn{1}{c}{Models} &
  \multicolumn{1}{c}{$\epsilon_r(u,\hat{u})$} &
  \multicolumn{1}{c}{MAE} &
  \\ \midrule
  
Entire state space learning \cite{krishnapriyan2021characterizing}  & $9.60 \times 10^{-1}$& $6.84 \times 10^{-1}$ \\

Sequence-to-sequence learning ($\Delta = 0.05$) \cite{krishnapriyan2021characterizing}  & $2.81 \times 10^{-2}$& $1.17\times 10^{-2}$ \\

Sequence-to-sequence learning ($\Delta = 0.1$) \cite{krishnapriyan2021characterizing}  & $2.69 \times 10^{-2}$& $1.28 \times 10^{-2}$ \\

Proposed method (learning the entire state space) & $\boldsymbol{9.326 \times 10^{-4}}$ & $\boldsymbol{6.312 \times 10^{-4}}$ \\
\end{tabular}}
\end{table*}

\section{Conclusion}\label{sec:Conclusion}
Backpropagation is central to the training of neural networks. Essentially, it is an efficient method for calculating the gradient of an objective function with respect to the weights of a neural network model. In conventional machine learning applications, rich sets of data are used to train deep neural network models in a supervised learning fashion.  However, when learning the solution of a differential equation, such data is either unavailable or unfeasible to gather. Therefore, an alternative approach to training neural network models for scientific applications is to incorporate governing equations of the physics problem at hand, which are generally partial differential equations (PDEs). However, high-order differential operators are sources for amplification of learning complexity in physics informed/constrained neural networks. Output from a neural network model during early stages of training contains noise or perturbations. Approximating physics from noisy output leads to more noisy physics loss, which consequently corrupts the back-propagated gradients and impede convergence. We have shown this issue by injecting perturbations in the output obtained from a neural network model after training for a small number of epochs, which have produced high levels of noise in a structured loss function that is composed of differential operators. To quantify the impact of noise, we juxtaposed the distribution of back-propagated gradients before and after the noise injection. To mitigate this training difficulty, we introduced auxiliary flux parameters to reduce the order of a given PDE and relax stringent smoothness requirement on its solution. In doing so, we avoid calculating high order derivatives. We then carefully formulated an unconstrained optimization problem that properly embeds the relevant physics in our objective function while tightly enforcing the boundary constraints and adaptively focusing on regions of higher gradients 
that are difficult to learn. We applied our methodology to various challenging PDE problems and demonstrated orders of magnitude improvements over other published results. 

\clearpage
\bibliographystyle{elsarticle-num-names} 
\bibliography{citations}

\begin{IEEEbiography}{Shamsulhaq Basir}
received a B.Sc. degree in aerospace engineering with a double major in civil engineering from the Middle East Technical University in Ankara, Turkey. He is currently a PhD candidate in the mechanical and material science department at the University of Pittsburgh. His research interests include deep learning, scientific machine learning, domain decomposition, computational fluid dynamics and high performance computing. 
\end{IEEEbiography}

\begin{IEEEbiography}{Inanc Senocak}{\space}is an associate professor of mechanical engineering and a William Kepler Whiteford Faculty Fellow at the University of Pittsburgh. He obtained his PhD degree in aerospace engineering from the University of Florida and his B.Sc. degree in mechanical engineering from the Middle East Technical University in Ankara, Turkey. He worked as a postdoctoral researcher at the Stanford University and the Los Alamos National Laboratory prior to starting his faculty career at Boise State University in 2007. He is a fellow of the American Society of Mechanical Engineers (ASME), an associate fellow of the American Institute of Aeronautics and Astronautics (AIAA), and a past recipient of a CAREER Award from the National Science Foundation. 
\end{IEEEbiography}

\end{document}